\documentclass[letterpaper, 10 pt, journal, twoside]{IEEEtran}

\usepackage{times}
\usepackage{multicol}
\usepackage[bookmarks=true]{hyperref}
\usepackage[caption=false]{subfig}
\usepackage{float, mathtools, amssymb, csquotes, xcolor, pgfplots, svg}
\pgfplotsset{compat=1.18}

\newcommand{\networkName}{MotionPerceiver}

\begin{document}

\title{\networkName{}: Real-Time Occupancy Forecasting for Embedded Systems}

\author{Bryce Ferenczi$^{1}$\textsuperscript{\textdagger}, Michael Burke$^{1}$, Tom Drummond$^{1,2}$
\thanks{Manuscript received: September, 25, 2023; Revised December, 26, 2023; Accepted January, 18, 2024.}
\thanks{This paper was recommended for publication by Editor Cesar Cadena Lerma upon evaluation of the Associate Editor and Reviewers' comments.
This work was supported by an Australian Government Research Training Program (RTP) Scholarship}
\thanks{$^{1}$Department of Electrical and Computer Systems Engineering, Monash University, Australia}%
\thanks{$^{2}$Department of Computing and Information Systems, Melbourne University, Australia}%
\thanks{\textsuperscript{\textdagger} Corresponding Author: {\tt\small bryce.ferenczi@monash.edu}}
\thanks{Digital Object Identifier (DOI): see top of this page.}
\thanks{\textcopyright 2024 IEEE.  Personal use of this material is permitted.  Permission from IEEE must be obtained for all other uses, in any current or future media, including reprinting/republishing this material for advertising or promotional purposes, creating new collective works, for resale or redistribution to servers or lists, or reuse of any copyrighted component of this work in other works.}
}

\markboth{IEEE Robotics and Automation Letters. Preprint Version. Accepted January, 2024}
{Ferenczi \MakeLowercase{\textit{et al.}}: \networkName} 

\definecolor{darkred}{RGB}{165,0,38}
\definecolor{lightred}{RGB}{215,48,39}
\definecolor{darkorange}{RGB}{244,109,67}
\definecolor{lightorange}{RGB}{253,174,97}
\definecolor{lightblue}{RGB}{116,173,209}
\definecolor{mediumblue}{RGB}{69,117,180}
\definecolor{darkblue}{RGB}{49,54,149}
\definecolor{my-green}{RGB}{0,168,0}

\maketitle

\begin{abstract}
This work introduces a novel and adaptable architecture designed for real-time occupancy forecasting that outperforms existing state-of-the-art models on the Waymo Open Motion Dataset in Soft IOU. The proposed model uses recursive latent state estimation with learned transformer-based functions to effectively update and evolve the state. This enables highly efficient real-time inference on embedded systems, as profiled on an Nvidia Xavier AGX. Our model, \networkName{}, achieves this by encoding a scene into a latent state that evolves in time through self-attention mechanisms. Additionally, it incorporates relevant scene observations, such as traffic signals, road topology and agent detections, through cross-attention mechanisms. This forms an efficient data-streaming architecture, that contrasts with the expensive, fixed-sequence input common in existing models. The architecture also offers the distinct advantage of generating occupancy predictions through localized querying based on a point-of-interest, as opposed to generating fixed-size occupancy images that render potentially irrelevant regions.
\end{abstract}


\IEEEpeerreviewmaketitle

\section{Introduction}
\IEEEPARstart{M}{otion} forecasting is a crucial step in trajectory planning for autonomous vehicles, integral to optimal decision making and preventing collisions with other agents in a dynamic environment. This task, however, is fraught with complexities, owing to the multifaceted factors that influence the trajectories of other agents. These factors encompass the environmental context, per-agent goals, navigable area and social interactions between agents. Compounding this challenge is the heterogeneity of the data structures derived from the sensor suite of autonomous vehicles. This data spans a spectrum of temporal attributes, ranging from time-sensitive (e.g. the position of other agents or traffic signals) to time-invariant (e.g. lane markings or pedestrian crossings). This data diversity creates challenges around representation, as it can manifest in sparse and concise forms, resembling a singular point in the environment, or more intricate forms, exemplified by lane markings encoded as directional splines. Furthermore, the volume of data within a scene varies significantly, influenced by instance-based features, such as agents and lane markings. Practical use within autonomous systems introduces a number of additional requirements: real-time capability, efficient query processes for downstream algorithms, and uncertainty estimation to enable risk-based motion planning.

\begin{figure}
    \centering
    \begin{tabular}{ccc}
        \subfloat[Last Observation ($t=0s$)]{\includegraphics[width=1in]{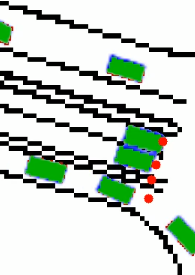}} &
        \subfloat[Before Next Observation ($t=0.4s$)]{\includegraphics[width=1in]{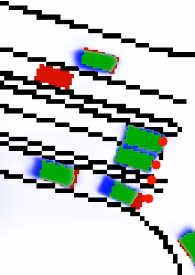}} &
        \subfloat[Next Observation ($t=0.5s$)]{\includegraphics[width=1in]{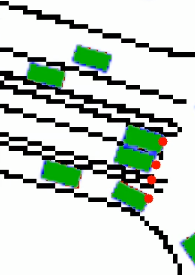}} \\
    \end{tabular}
    \caption{\networkName{} implements recursive state estimation for motion forecasting. Here, an initial scene observation ($a$) at $t=0s$  is forecast ($b$)  $0.4s$ into the future. This learned prediction begins to accumulate error and uncertainty for targets in motion, and shows no occupancy for an agent that was not initially detected. Frame ($c$) shows the predicted occupancy after a scene observation has applied an update to the latent state at $t=0.5s$. Agent occupancy is refined, without explicit data association, and a new agent added to the state. Images are color coded \textcolor{my-green}{green} $\rightarrow$ true positive (occupancy prediction $>0.5$), \textcolor{blue}{blue} $\rightarrow$ false positive, \textcolor{red}{red} $\rightarrow$ false negative, black $\rightarrow$ rasterized road graph, \textcolor{red}{red dots} $\rightarrow$ traffic signals.} 
    \label{fig:updatestep}
\end{figure}

This paper introduces \networkName{}, a real-time motion forecasting model explicitly designed to handle these representation challenges, seamlessly integrating information from a broad range of sensors. At its core, \networkName{} models a scene as a latent state that continually evolves in time with updates derived from latent representations of scene observations as they become available. Importantly, this architecture uses a design cue from Perceiver-IO \cite{PercieverIO}, using fixed latent state dimensions to facilitate deterministic memory and computational cost. \networkName{} leverages a learned time evolution function to predict the future latent state, rooted in multi-head self-attention mechanisms to capture dynamic interactions among features embedded in the latent state, such as vehicles and pedestrians. In order to assimilate incoming data, such as agent positions and traffic signals, we employ cross-attention to facilitate information transfer to the projected latent state. This scheme can be considered analogous to a recursive state estimation process, improving the predicted latent scene representation as novel information emerges (Fig. \ref{fig:updatestep}). Predicting occupancy is performed by querying the latent state at the desired time, using an encoding of the desired query position. For a comprehensive occupancy prediction of the scene, a grid of positions can be used to generate a rasterized birds-eye-view image.


Motion forecasting necessitates operating at a rate surpassing real-time to incorporate new observations of the scene. To efficiently process a causal signal, it is preferable to only process new incoming information, rather than redundantly reprocessing old data. Algorithms that use a fixed-sequence for input are inefficient since they begin inference from scratch and reprocess historical information. Unfortunately, this is the dominant paradigm used by the majority of motion forecasting architectures (\cite{ChauffeurNet,DenseTNT,HOPE,VectorFlow,STOPNet,MotionCNN,STrajNet,OccupancyFlow,SceneTransformer,LookAround,SoPhie,Trajectron++,MultiPath++,AgentFormer}). A more streamlined approach would be to augment the prior motion forecast with new information. Furthermore, scalability is also paramount, to cope with a large numbers of agents. Solutions that scale super-linearly or are not parallelisable are impractical for real-world deployment. \networkName{}'s latent state prediction-update scheme is designed with a computationally-deterministic data streaming paradigm in mind. This allows the previous computational investment in building the latent state representation to be retained, while seamlessly incorporating new measurements.

Finally, robustness to noisy or incomplete data is essential for real-world deployment where occlusions and sensor blind spots are common. Sequence-based models commonly use agent tracklets as input, which can be corrupted during occlusions or a faulty tracking algorithm that outputs switched ids. \networkName{}'s latent state update does not use temporal data association at the input, and is robust to this problem.
In summary, the core contributions of this work are:
\begin{itemize}
\item A novel transformer-based architecture that uses recursive state estimation to forecast occupancy using agent detections and seamlessly integrates contextual information.
\item The first data-streaming based motion forecasting model that reports real-time inference on an edge device.
\end{itemize}
This model outperforms existing models in predicted occupancy (Soft IoU) on the Waymo Open Motion dataset.

\section{Related Work}
\textbf{Motion forecasting} is the task of predicting the future position of agents in a scene. There are two reference frames which are commonly used, agent-centric and scene-centric. Agent-centric methods \cite{MotionCNN} transform the scene into each agent's point-of-view for prediction. This paradigm can become prohibitively expensive due to inference cost scaling per agent in the scene. Scene-centric methods holistically encode the entire scene and jointly predict motion of all agents. This is the dominant architecture in the literature, and the paradigm followed by \networkName{}. A variety of scene-encoding architectures have been explored in this category including: point-pillars \cite{STOPNet,PointPillars,OccupancyFlow}, fully-convolutional \cite{MotionCNN,LookAround}, transformer \cite{STrajNet}, graph \cite{VectorFlow} and hybrid methods \cite{HOPE}.

\textbf{Trajectory forecasting} involves predicting continuous splines that represent the future trajectory for each agent. This can take the form of a probabilistic set, representing a multi-modal decision distribution of the target \cite{DenseTNT,MotionCNN,Trajectron++,MultiPath++}. A variety of higher-level architectural designs have been explored to accomplish this task. State-of-the-art trajectory planners commonly follow a paradigm of finding a set of likely goals, and generate feasible trajectories towards these \cite{DenseTNT,GANet}. Anchors are also used as a foundation for multi-modal prediction forecasting. Anchors can be derived statistically from the dataset \cite{MultiPath} or learned during training \cite{MultiPath++}. Other methods forecast future trajectories without using anchors or goals as grounding \cite{MotionCNN,SceneTransformer,Trajectron++,AgentFormer}. An important formalisation is to predict control inputs for the motion model of the agent to prevent physically infeasible predictions \cite{Trajectron++,MultiPath++}. The weakness of per-agent trajectory forecasting is the high complexity this formulation can elicit in larger, crowded scenes. A path-planning algorithm has to parse, optimize and validate a multi-modal trajectory distribution for each agent, adding a large number of constraints.

\textbf{Agent-based Occupancy forecasting} on the other-hand, is the goal of predicting if a point in space will be occupied by an agent in the future. This paradigm is more amenable to path planning algorithms as they can check whether a position in the navigable area is occupied with a single query. This formulation, in isolation, erases the identities of agents. Methods which also predict the flow of occupancy over time \cite{HOPE,VectorFlow,STOPNet,STrajNet,OccupancyFlow,LookAround} enable retracing from where occupancy has originated, and the agent responsible. Occupancy forecasting implicitly permits multi-modal predictions as the occupancy attributed to a particular agent is able to spread beyond the real size of the agent as its position becomes more uncertain further in the future. Occupancy prediction is performed by \networkName{} due to its ease-of-use in downstream planning or trajectory optimisation algorithms.

\textbf{Sensor-based Occupancy Forecasting} estimates birds-eye-view (BEV) or volumetric occupancy of a scene using time-of-flight sensors such as Lidar and/or Radar. This contrasts with the object conditioned occupancy forecasting considered by \networkName{}. Intrinsic challenges to this task are sparse-to-dense reconstruction, and reasoning around sensor-occluded areas of the scene. A common approach is to predict occupancy from a Lidar point-cloud observation, forecast to the next observation's space-time location, then enforce consistency with the next point-cloud \cite{PointcloudForecastingAsProxy,LidarCompletionForecasting,RTXSelfSupervisedOccupancy}. Radar introduces additional challenges due to various noise modalities such as reflection and receiver saturation. Despite these challenges, its longer range, penetration and lower cost make it an ideal candidate for use in autonomous vehicles. Hence, \cite{RadarOccupancyPrediction} develop a model to predict occupancy from Radar, using accompanying Lidar supervision for training.

\textbf{Including scene context} is an important capability for forecasting models, especially for predicting over longer time horizons. Context can be invariant over time -- this typically includes topographical information such as lane markings, potholes or signage. Time-sensitive context includes dynamic entities such as traffic signalling equipment and other agents in the environment. Motion forecasting models within autonomous driving settings predominantly use lane markings as context \cite{DenseTNT,VectorFlow,LaneGCN,STrajNet,GANet}. Lane marking information, represented as directed splines, are rarely used as raw input to a model. Preprocessing into a more amenable representation is needed. Commonly, lane markings are rasterized into an image that can be processed by a CNN for feature extraction \cite{STrajNet,MotionCNN}. Alternatively, VectorNet \cite{VectorNet} uses a hierarchical graph neural network for encoding. Individual splines are initially processed into features of fixed dimension, then global interactions are modeled between these features, resulting in a concise representation of each spline as a feature of fixed dimension. A smaller subset of models include traffic signals \cite{HOPE,STOPNet,OccupancyFlow,SceneTransformer}, but this is likely attributed to some popular datasets \cite{Argoverse2,Interaction} not containing high quality traffic signal information and its limited quantitative effect on performance. \networkName{} is explicitly designed to be extendable and integrate a variety of features that may influence agent dynamics, which is important as new contextual features become available in driving datasets.

The vast majority of forecasting models consume a \textbf{fixed sequence} of observations for inference due to the simplicity of capturing spatio-temporal features as a single input. The most common method is concatenating the history of an agent into a feature-vector \cite{VectorNet,VectorFlow,AgentFormer}. A small number of methods accumulate timestamped positions of agents in a point-cloud \cite{STOPNet,OccupancyFlow}. In general, these sequence based models have the disadvantage that inference is started anew for each forecast when a new scene observation is captured. Furthermore, accurate tracking and data association methods are required to build the history for each agent (a tracklet). Motion forecasting performance could be potentially degraded if malformed tracklets are introduced by tracking algorithm errors. \networkName{} uses a streaming based approach that avoids repetition of inference and use of tracklets. Here, we update our forecast as new observations come in, based on our previous forecast.

\section{Preliminaries}
\subsection{Transformer attention}
The multi-head attention function of the \textbf{transformer} \cite{AttnIsAllYouNeed}, denoted as $\mathrm{MHA}$, has become pervasive in deep learning.  The output of each attention head, $\mathbf{H}_i,i\in\{1,\ldots, h\}$, is a weighted sum of $\mathbf{V}_i$, using learned correlations between queries $\mathbf{Q}_i$ and keys $\mathbf{K}_i$,
\begin{equation}
\mathbf{H}_i=\mathrm{Attn}(\mathbf{Q}W_i^q, \mathbf{K}W_i^k, \mathbf{V}W_i^v)=\mathrm{softmax}(\frac{\mathbf{Q}_i \mathbf{K}_i^T}{\sqrt{d_k}})\mathbf{V}_i,
\label{eq:attn}
\end{equation}
where each attention head uses a different projection of inputs $\mathbf{Q}$, $\mathbf{K}$ and $\mathbf{V}$, using learned parameters $W_i^q\in \mathbb{R}^{d\times d_q}$,$W_i^k\in \mathbb{R}^{d\times d_q}$,$W_i^v\in \mathbb{R}^{d\times d_v}$. For clarity, $d$ is the dimension of the model output, $d_q$ is the dimension of $\mathbf{K}$ and $d_v$ is the dimension of $\mathbf{V}$. The result from each head is linearly combined using learned parameter $W^o\in \mathbb{R}^{hd_v \times d}$
\begin{equation}
\mathbf{\hat{H}} = \mathrm{MHA}(\mathbf{Q},\mathbf{K},\mathbf{V}) = \mathrm{concat}(\mathbf{H}_1,\ldots,\mathbf{H}_h)W^o.
\label{eq:tformer}
\end{equation}
A typical transformer block includes a residual pass through a multi-layer perceptron ($\mathrm{mlp}$), to produce final output $\mathbf{H'}$,
\begin{equation}
\mathbf{H'} = \mathrm{mlp}(\mathbf{\hat{H}}) + \mathbf{\hat{H}}.
\label{eq:tformer-mlp}
\end{equation}
We use transformer blocks as the foundation of each \networkName{} function to model the interactions between features in the latent state and observations from the scene.

\subsection{Sinusoidal Position Encoding}
Transformer models are often paired with a positional encoding scheme, enabling better identification of distance-based correlations between tokens. Common positional features in motion forecasting include coordinates $(x,y)\rightarrow\mathbf{p}\in\mathbb{R}^2$ and pose $(x,y,\theta)\rightarrow\mathbf{\hat{p}}\in\mathbb{R}^3$, where $x$ and $y$ denote position, and $\theta$ represents heading in a reference frame. The sinusoidal position encoding scheme uses a series of $f_n$ evenly spaced frequency components $f_i$, from $f_{min}$ to $f_{max}$. To encode a scalar position $p$, each element is calculated 
\begin{equation}
    \mathcal{P}_i(p) = 
    \begin{cases}
    \sin(f_i \pi p)\quad i<f_n\\
    \cos(f_i \pi p)\quad else,\\
    \end{cases}
\label{eq:sinusoid_elem}
\end{equation}
and concatenated to form the sinusoidal encoding vector, 
\begin{equation}
    \mathcal{P}(p) = \mathrm{concat}(\mathcal{P}_1(p),\dots,\mathcal{P}_{2\cdot f_n}(p))
\label{eq:sinusoid_enc}
\end{equation}
For a vector, $\mathbf{\hat{p}}$, this procedure is repeated for each element,
\begin{equation}
    \mathcal{P}(\mathbf{\hat{p}})=\mathrm{concat}(\mathcal{P}(\mathbf{\hat{p}}_x), \mathcal{P}(\mathbf{\hat{p}}_y), \mathcal{P}(\mathbf{\hat{p}}_{\theta})).
\label{eq:sinusiod_vec}
\end{equation}

\begin{figure*}[hbt]
    \centering
    \includegraphics[width=6in]{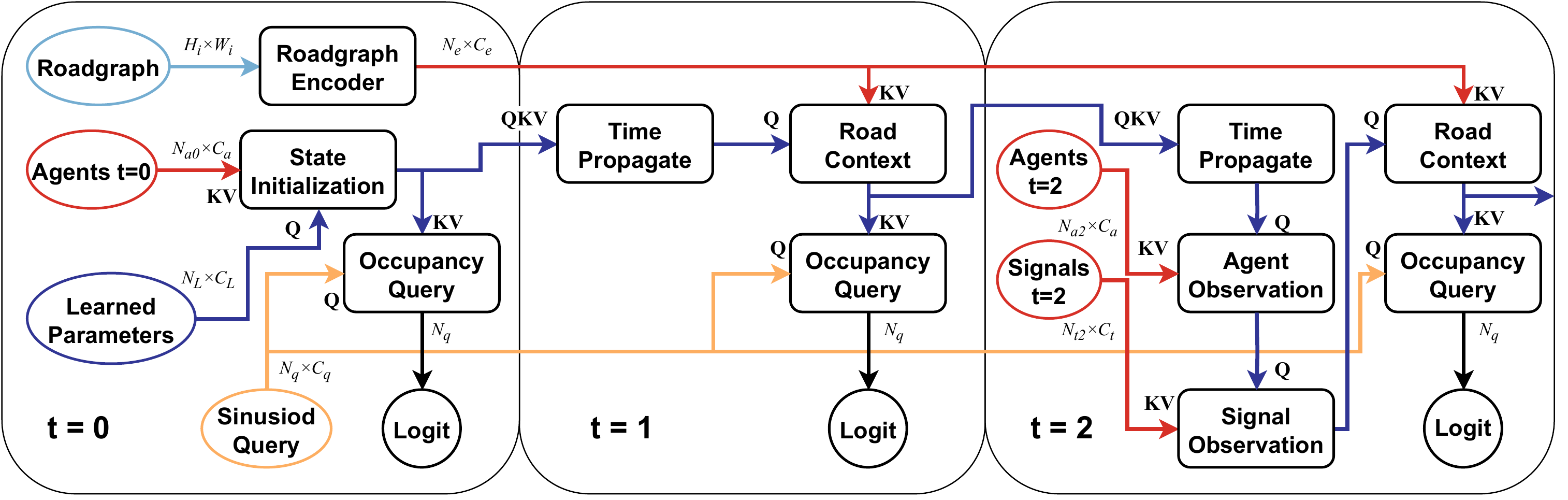}
    \caption{An illustration of the use of \networkName{}'s architecture for real-time occupancy prediction. At $t=0$, the latent state is initialised with the first observation of agents in the scene. The evolution of this latent state is shown using \textcolor{darkblue}{dark blue} arrows. Tokenized observations from the scene (\textcolor{lightred}{light red}) are queried by the latent state for information. Rasterized road-graph context (\textcolor{lightblue}{light blue}) can be encoded once and provide contextual information at each time-step. The latent state can be queried with a position (\textcolor{lightorange}{light orange}) to receive an estimate of occupancy probability at each time-step. When there is no observation information ($t=1$), the latent state is simply propagated forward in time and updated with road-graph context. This is the operation used for forecasting future occupancy or interpolation between observations. At time-steps when scene observations are available ($t=2$), the latent state queries the observation for information, reducing accumulated errors and adding newly observed agents. Dimensions $N_x\times C_x$ describes the number of tokens $N_x$ and channels per token $C_x$. Additional inference diagrams can be found at \href{https://sites.google.com/monash.edu/motionperceiver}{https://sites.google.com/monash.edu/motionperceiver}.}
    \label{fig:model_arch}
\end{figure*}

\section{\networkName{}}
\subsection{Problem formulation}
Our goal is to make predictions of the future occupancy within a region of interest, conditioned on observations that include the positions, orientations and velocities of other agents in a scene, traffic signalling information, and other contextual information such as road lane markings. We assume that agent id is not available, and the number of agents observed can vary over time, due to occlusions, or agents entering and leaving the scene. We consider a streaming data paradigm, where sensors are providing these observations periodically, in real time.

\subsection{Tokenizing Observations}
A key advantage of the \networkName{} architecture is adaptability to ingest diverse data sources, needing only a reasonable strategy to encode information into a set of $N_o$ tokens with dimension $C_o$,  $\mathbf{I_t}= \{I_t^0,\ldots,I_t^{N_o}\},I_t^n\in\mathbb{R}^{C_o}$. In this section, we outline the method used for transforming raw agent, traffic signal and road-graph observations into a form \networkName{} can consume. Using a scene region-of-interest of $160\times 160m$, position is normalized to $[-1,1]$ and sinusiodally encoded with (\ref{eq:sinusiod_vec}). We use $f_{min}=1Hz$, $f_{max}=320Hz$ and $f_n$ is specified per-feature, dependent on desired fidelity. 

A number of properties from the agent observation are used to construct the tokenized representation used by \networkName{}. Observed pose $\mathbf{\hat{p}}$, is sinusoidally encoded with $f_n=64$. Vehicle dimensions, $(h,w)$, are incorporated to rasterize a correctly sized occupancy mask. Rate-of-change $\mathbf{\dot{\hat{p}}}$, enables identification of vehicles in motion with one observation. These features are concatenated to produce a vector representation of each visible agent $A=(\mathcal{P}(\mathbf{\hat{p}}), \mathbf{\dot{\hat{p}}}, h, w)$ with dimension $C_a$.  Hence, for inference, a variable number of agents $N_a$, observed at time $t$, are transformed into a set of tokens $\mathbf{A}_t=\{A^1,\ldots,A^{N_a}\}_t$.

To construct a traffic signal token, observed position $\mathbf{p}$ is sinusoidally encoded with $f_n=32$ and concatenated with a one-hot encoding of the signal type $\mathcal{T}$ \footnote{Examples of signal types include green straight and red turn.}, producing feature vector $T=(\mathcal{P}(\mathbf{p}), \mathcal{T})$ of dimension $C_t$. For $N_t$ traffic signals observed at time $t$, we create a set $\mathbf{T}_t=\{T^1,\ldots,T^{N_t}\}_t$.

A rasterization strategy is used to tokenize road-graph information for \networkName{}. Lane markings are rasterized as a $H_i\times W_i$ image and passed through a convolutional neural network (CNN), creating a $H_p\times W_p$ feature image. Each pixel value $\hat{R}$, is a high-dimensional representation of a section of the map. This is concatenated with a sinusoidal encoding ($f_n=16$) of its location $\mathbf{p}$, creating a feature vector $R=(\hat{R}, \mathcal{P}(\mathbf{p}))$ of dimension $C_e$. Hence, from a variable number the lane markings in a scene, we create a fixed set of tokens $\mathbf{R}=\{R^1,\dots,R^{N_e}\}$ where $N_e$ is the number of pixels $H_p\times W_p$. This decouples run-time cost from the density of road-graph content in the scene\footnote{Creating a rasterized image of splines is relatively inexpensive.}.

\subsection{Recursive State Estimation Using Transformers}

\begin{figure}
    \centering
    \includegraphics[width=3.3in]{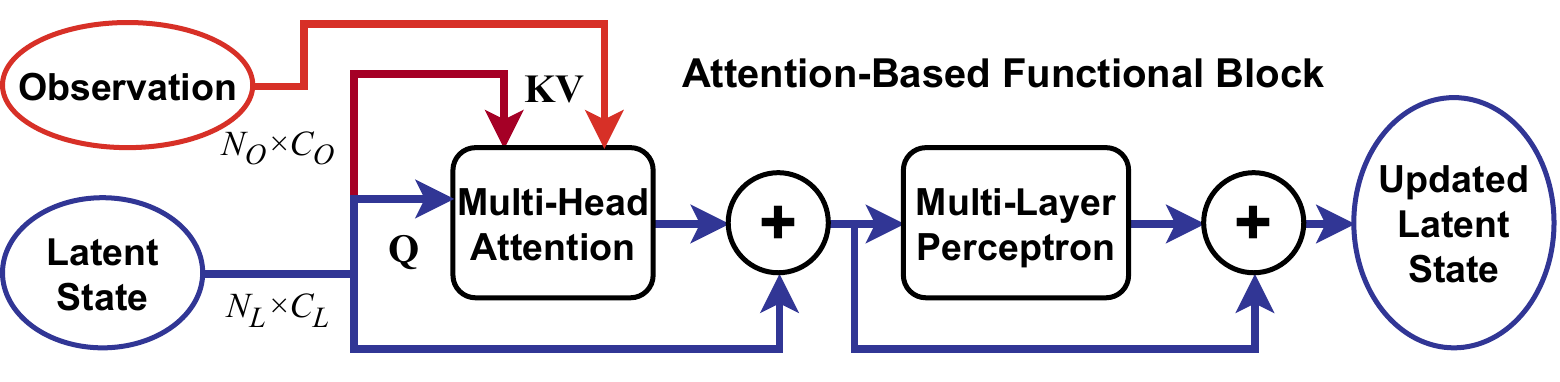}
    \caption{Self- and cross-attention is used to apply changes to the latent state. The latent state (\textcolor{darkblue}{dark blue}), is always used as the query. Self-attention sources the key-value from the latent state (\textcolor{darkred}{dark red}), communicating information between the $N_L$ variables in the latent state. Cross-attention uses the observation as the key-value (\textcolor{lightred}{light red}) for the latent state to query and transfer information from the observation to the latent state.}
    \label{fig:attnFunc}
\end{figure}

\networkName{}, shown in Fig. \ref{fig:model_arch}, models agents navigating a dynamic social environment. We learn how a representation of the scene, described as a set of latent variables $\mathbf{S}=\{S_1,\ldots,S_{N_L}\}, S_n\in\mathbb{R}^{C_L}$, evolves in time (shown in the \textcolor{darkblue}{dark blue} path of Fig. \ref{fig:model_arch}). Self- and cross-attention based functions allow the model to capture and express social dynamics between the agents in the latent state and context when predicting the future.

To initialize the latent state, $\mathbf{S}$, a set of learned parameters, query the first observation of agents, $\mathbf{A}_0$, with cross-attention. This is followed by $6$ self-attention applications. This forms our $StateInitialization$ module, depicted in Fig. \ref{fig:model_arch}. From here, we begin performing recursive state estimation with \networkName{}.

The time evolution process modeled as learned function $\mathcal{F}$, only depends on the previous state $\mathbf{S_{t-1|\cdot}}$, to predict the next state $\mathbf{S_{t|t-1}}$, akin to a Markovian dynamics model, 
\begin{equation}
\mathbf{S_{t|t-1}} = \mathcal{F}(\mathbf{S_{t-1|\cdot}}).
\label{eq:forward}
\end{equation}
Here, we adopt Kalman filter notation, $\mathbf{S_{t|t-1}}$, to represent a state at time $t$, given observations at time $t-1$. The latent state $\mathbf{S_{t-1|\cdot}}$ is used for the arguments in (\ref{eq:tformer}),
\begin{equation}
\mathbf{Q} = \mathbf{S_{t-1|\cdot}},\; \mathbf{K} = \mathbf{S_{t-1|\cdot}},\; \mathbf{V} = \mathbf{S_{t-1|\cdot}},
\label{eq:selfattnkqv}
\end{equation} 
creating a self-attention between the $N_L$ variables in the latent state. This process is visualized in the \textcolor{darkblue}{dark blue} and \textcolor{darkred}{dark red} path(s) of Fig. \ref{fig:attnFunc} where $\mathbf{S_{t-1|\cdot}}$ is the ``Latent State" input. This self-attention function is repeated $6$ times in a block to create the $TimePropagate$ module in Fig. \ref{fig:model_arch}.

To transfer information from tokenized observation $\mathbf{I_t}$ to the latent state, a learned update function $\mathcal{U}$ is used \textbf{after} the state has been propagated to the time frame when the information was captured,
\begin{equation}
\mathbf{S_{t|t}} = \mathcal{U}(\mathcal{F}(\mathbf{S_{t-1|\cdot}}), \mathbf{I_t}) = \mathcal{U}(\mathbf{S_{t|t-1}}, \mathbf{I_t}).
\label{eq:update}
\end{equation}
Here, the latent variables of $\mathbf{S_{t|t-1}}$ query the key-value pairs derived from $\mathbf{I_t}$ using (\ref{eq:tformer}), 
\begin{equation}
\mathbf{Q} = \mathbf{S_{t|t-1}},\quad\mathbf{K} = \mathbf{I_t},\quad\mathbf{V} = \mathbf{I_t}.
\label{eq:crossattnkqv}
\end{equation}
This cross-attention process is depicted in Fig. \ref{fig:attnFunc} and follows the \textcolor{lightred}{light red} path for ``Observation" $\mathbf{I_t}$, and the \textcolor{darkblue}{dark blue} path for ``Latent State" $\mathbf{S_{t|t-1}}$. This is used in the $RoadContext$, $Agent Observation$ and $SignalObservation$ modules. An additional self-attention is applied after cross-attention in the $RoadContext$ and $SignalObservation$ modules.

The latent state can be queried with a set of positions $\mathbf{P}=\{\mathbf{p}_1,\ldots,\mathbf{p}_{N_q}\}$ to yield predicted occupancy logits, $\hat{o_t}$, given the current state of the scene $\mathbf{S_{t|\cdot}}$ with learned emission function $\mathcal{O}$,
\begin{equation}
\hat{o_t} = \mathcal{O}(\mathbf{P}, \mathbf{S}_{t|\cdot}) = Pr(\mathbf{P} | \mathbf{S_{t|\cdot}}).
\label{eq:output}
\end{equation}
Once again, this is performed using cross-attention (\ref{eq:tformer}),
\begin{equation}
\mathbf{Q} = \mathcal{P}(\mathbf{P}),\quad\mathbf{K} = \mathbf{S_{t|\cdot}},\quad\mathbf{V} = \mathbf{S_{t|\cdot}}.
\label{eq:outputQuery}
\end{equation}
A potential weakness of point-wise methods is missing near-field context. If greater accuracy is desired, a region of pixels can be queried and decoded with a small CNN (Conv Decode). Importantly, this does not nullify our ``local query" capability, we still query a patch rather than the full scene.

Intuitively, these operations can be considered analogous to a recursive Bayesian estimation filtering operation, with a learned latent dynamics model making future predictions over a state, and a learned update model refining the latent state with later observations. The attention mechanisms learn to introspect and update specific latent variables. An emission model generates a probabilistic occupancy map as a function of the latent state conditioned on a series of observations.

\subsection{Loss Function}
To predict pixel-wise occupancy, a focal loss \cite{focal} is used to address class imbalance between occupied and unoccupied pixels. Formally, focal loss $\mathcal{L}_f$, is defined for occupancy prediction logit $\hat{o}$ and ground truth binary occupancy $o$, where $\sigma$ is the sigmoid activation function and $\gamma$ is the focal weighting factor,
\begin{align}
\mathcal{F} &= \sigma(\hat{o})o + (1 - o)(1 -\sigma(\hat{o})), \\
\mathcal{L}_f &= -\log(\hat{o})(1 - \mathcal{F})^\gamma
\label{eq:focal}
\end{align}
An additional weighting factor $\alpha$ is used to weight positive samples,
\begin{equation}
\alpha_f = \alpha\hat{o} + (1 - \alpha)(1 - o)
\label{eq:alpha_w}
\end{equation}
and is uniformly averaged over all time-steps $N_{time}$ and occupancy pixels $N_{pixel}$ for final loss $\mathcal{L}$,
\begin{equation}
\mathcal{L} = \frac{1}{N_{time}N_{pixel}} \sum_{N_{time}}\sum_{N_{pixel}} \alpha_f\mathcal{L}_f.
\label{eq:ave_loss}
\end{equation}

\subsection{Uncertainty Calibration}
For an imbalanced binary classification task, focal loss parameters $\gamma$ and $\alpha$, can be adjusted to balance the false-negative and false-positive rate. For temporal prediction this is likely to be a function of prediction time. Finding the ideal focal parameters at every time step would entail an expensive hyper-parameter search, so we train with a fixed set of parameters, and calibrate the model output instead by applying a simple scaling function to $\hat{o}$, increasing the decay rate of predicted occupancy by a factor $\beta$,
\begin{equation}
\hat{o} = 
    \begin{cases}
        \beta \hat{o} & \text{if } \hat{o}<0 \\
        \hat{o} & \text{else }.
    \end{cases}
\label{eq:scale_func}
\end{equation}

\subsection{Online Inference}
\label{sec:online_inference}
A key design feature of \networkName{} is the real-time streaming-based architecture. To deploy this in an online application, the latent state that aligns with timing of the next anticipated observation is preserved. Retaining a single latent state entails a constant memory requirement of $N_L\times C_L$. The next scene observation update can be applied to the retained latent state, thereby facilitating the seamless resumption of the forecasting process. This is conceptually a more efficient paradigm compared to more common sequence-based methods. While recurrent models theoretically enable streaming, these models perform poorly in motion forecasting \cite{jia2022towards}, or are applied to latents or tokens produced by passing in fixed windows of information. To the best of our knowledge, there are no streaming-based architectures reported in the literature capable of matching fixed sequence models in performance.

If the sensor sample period and the requested forecasting period differ, more than one $TimePropagate$ function can be trained to evolve the latent state at different periods. This more computationally efficient for inference than a least-common-multiple $TimePropagate$ that needs to be applied multiple times to match the desired period.


\section{Experimental Results}

\begin{table*}[hbt]
    \centering
    \vspace{3mm}
    \caption{Architecture Ablation on Waymo Open Motion Validation Split}
    \begin{tabular}{|c|c|c|c|c|c|c|c|c|c|c|c|}
        \hline
        \textbf{Latent Dims}& \textbf{Uncertainty}& \multicolumn{3}{|c|}{\textbf{Input Data}}& \textbf{Conv}&  \textbf{Two}& \multicolumn{3}{|c|}{\textbf{Soft IoU}}& \textbf{AUC}& \textbf{\#}\\
        \textbf{(N x C)}& \textbf{Calibration}& \textbf{Agents}& \textbf{Signal}& \textbf{Road}& \textbf{Decode}& \textbf{Phase}& \textbf{+3s}& \textbf{+6s}& \textbf{Mean}& \textbf{Mean}& \textbf{Params}\\
        \hline
        128x256& \checkmark& \checkmark& -& -& -& -& 0.443& 0.320& 0.424& 0.677& 5.84M\\
        128x256& \checkmark& \checkmark& \checkmark& -& -& -& 0.452& 0.329& 0.433& 0.685& 6.57M\\
        128x256& \checkmark& \checkmark& -& \checkmark& -& -& 0.454& 0.328& 0.433& 0.684& 7.50M\\
        128x256& \checkmark& \checkmark& \checkmark& \checkmark& -& -& 0.464& 0.333& 0.442& 0.693& 7.56M\\
        \hline
        64x128& \checkmark& \checkmark& \checkmark& \checkmark& \checkmark& -& 0.414& 0.308& 0.400& 0.651& \textbf{2.27M}\\
        128x256& -& \checkmark& \checkmark& \checkmark& \checkmark& -& 0.405& 0.265& 0.382& 0.716& 7.66M\\
        128x256& \checkmark& \checkmark& \checkmark& \checkmark& \checkmark& -& 0.486& 0.351& 0.463& 0.710& 7.66M\\
        \hline
        128x256& -& \checkmark& \checkmark& \checkmark& \checkmark& \checkmark& 0.501& 0.334& 0.460& \textbf{0.778}& 10.7M\\
        128x256& \checkmark& \checkmark& \checkmark& \checkmark& \checkmark& \checkmark& \textbf{0.563}& \textbf{0.400}& \textbf{0.524}& 0.772& 10.7M\\
        \hline
    \end{tabular}
    \label{tab:waymo_abl}
\end{table*}

\subsection{Training Environment}
We train and evaluate on Waymo Open Motion Dataset \cite{WaymoMotion} (WOMD), which consists of $104k$ $9$ second sequences captured at $10$Hz. Each sequence contains data including agent pose, traffic light signals and road topography. Unless otherwise specified, past phase data is sampled at $2$Hz. We use the same inertial frame as the WOMD evaluation benchmark. All models are trained with PyTorch \cite{PyTorch}\footnote{Code is available at \href{https://github.com/5had3z/motion-perceiver}{https://github.com/5had3z/motion-perceiver}.}. We use the ADAMW optimizer with an initial learning rate of $1e^{-3}$ and fixed batch size between $48-64$, depending on GPU memory available. A polynomial learning rate schedule is adopted with decay power of $0.9$ and epoch target of $75$. For occupancy focal loss (\ref{eq:ave_loss}), $\alpha=0.75$ and $\gamma=2$ are used. To generate an occupancy prediction, the latent state is queried with a evenly spaced $80\times 80m$ grid of $256\times 256px$, matching the ground truth occupancy mask. To conserve GPU memory, timesteps to query and apply the loss are sparsely sampled between $0$ and $6$ seconds. Random sampling mitigates spurious prediction artefacts that may arise on unsampled timesteps if only a consistent set is used. Additionally, we detach gradients between each timestep. This not only speeds up training, but also enforces the Markov assumption (\ref{eq:forward}).

\subsection{Performance metrics}
Performance in occupancy forecasting is characterized using Area Under Curve (AUC) and Soft Intersection-over-Union (Soft IoU). AUC, the area underneath the receiver operating characteristic (ROC) curve, is calculated by sampling the true and false positive rates at positive sample thresholds between 0 and 1. Soft IoU is defined as
\begin{equation}
Soft IoU=\frac{\sum_{N_{pixel}}o\hat{o}}{\sum_{N_{pixel}}(o+\hat{o}-o\hat{o})}.
\label{eq:soft_iou}
\end{equation}

\pgfplotsset{
    legend image with text/.style={
        legend image code/.code={%
            \node[anchor=center] at (0.3cm,0cm) {#1};
        }
    },
}
\begin{figure}
\vspace{1mm}
\begin{tikzpicture}
\begin{axis}[
    xlabel={Future (Seconds)},
    ylabel={Soft IoU / AUC},
    xmin=1, xmax=8,
    ymin=0, ymax=1,
    xtick={1,2,3,4,5,6,7,8},
    ytick={0,0.2,0.4,0.6,0.8,1.0},
    legend columns=2,
    legend pos=south west,
    legend style={
        font=\footnotesize,
        legend cell align=left,
    },
    ymajorgrids=true,
    grid style=dashed,
    every axis plot/.append style={thick}
]
\addlegendimage{legend image with text=Soft IoU}
\addlegendentry{}
\addlegendimage{legend image with text=AUC}
\addlegendentry{}

\addplot[color=lightred,mark=square]
    coordinates {
    (1,0.803849852587803)(2,0.670915936591761)(3,0.563138559238765)(4,0.485892484339422)
    (5,0.436548265239887)(6,0.400130386616912)(7,0.370788019619683)(8,0.347118046528138)
    };
    \addlegendentry{}
\addplot[color=lightred,mark=diamond,]
    coordinates {
    (1,0.96879389393625)(2,0.912738159912016)(3,0.834706507972098)(4,0.75648090376107)
    (5,0.697978321161206)(6,0.65073196486954)(7,0.60704332550378)(8,0.573622450017653)
    };
    \addlegendentry{Two Phase}
    
\addplot[color=darkorange,mark=square,]
    coordinates {
    (1,0.723771416177585)(2,0.553212290126293)(3,0.455953518200135)(4,0.396062863885035)
    (5,0.360510735264619)(6,0.33503050081698)(7,0.311422416539061)(8,0.290574728967027)
    };
    \addlegendentry{}
\addplot[color=darkorange,mark=diamond,]
    coordinates {
    (1,0.939143610877269)(2,0.83343713012276)(3,0.73744182478341)(4,0.662013162037076)
    (5,0.611234281971636)(6,0.571968094321463)(7,0.531032020117484)(8,0.49953994891893)
    };
    \addlegendentry{All Context}

\addplot[color=mediumblue,mark=square,]
    coordinates {
    (1,0.71459777077021)(2,0.540786674477805)(3,0.443144265779195)(4,0.384383011902757)
    (5,0.347730991782577)(6,0.319972894138)(7,0.292599886010963)(8,0.266943928321012)
    };
    \addlegendentry{}
\addplot[color=mediumblue,mark=diamond,]
    coordinates {
    (1,0.935918496722225)(2,0.82470602536234)(3,0.724885409437484)(4,0.648106893601617)
    (5,0.59645662840101)(6,0.555147994162193)(7,0.511752410899333)(8,0.475539473657024)
    };
    \addlegendentry{No Context}
\end{axis}
\end{tikzpicture}
\caption{Soft IoU and AUC at evaluation waypoints on Waymo Open Motion Validation split. Inclusion of contextual features (roadgraph + traffic signals) has a greater effect at later waypoints. Two phase prediction specialization improves performance across the whole sequence.}
\label{fig:iou_auc_perf}
\end{figure}
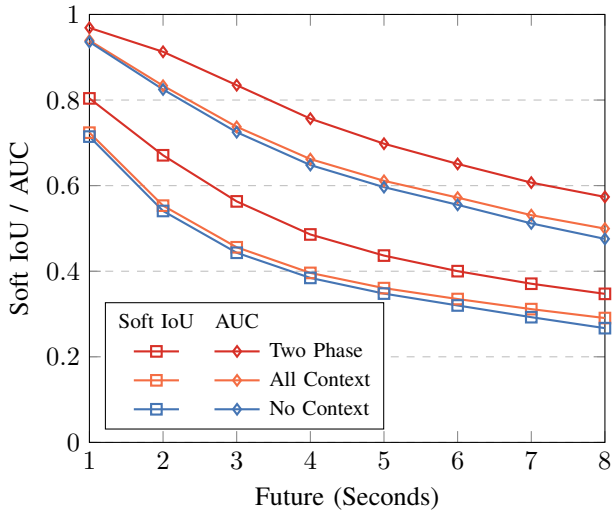

\subsection{General Insights}
We report architectural and feature ablations of \networkName{} in Table \ref{tab:waymo_abl}, investigating the effects of latent state dimension sizes, uncertainty calibration and adding scene context. We also explore the effects of ``Two Phase" training, discussed further in Section \ref{subs:comparison}, which uses different $TimePropagate$ modules for the past and future phase. For evaluation, mean Soft IoU and AUC are computed at $1s$ waypoints after the present. In general, \networkName{} is conservative at predicting occupancy, overestimating the probability of occupancy when it is unlikely. Hence, applying uncertainty calibration (\ref{eq:scale_func}) with $\beta=2$ results in improved Soft IOU and slight degradation in AUC (Uncertainty Calibration, Table \ref{tab:waymo_abl}). We also note incorporating contextual information has greater benefit over extended temporal horizons (Fig. \ref{fig:iou_auc_perf}). Decoding the occupancy query output with a two layer CNN instead of an MLP also shows a modest performance improvement (Conv Decode, Table. \ref{tab:waymo_abl}). Hence, a minor trade off between accuracy and latency is available.

\begin{figure}
    \centering
    \begin{tabular}{ccc}
         \subfloat[+0s]{\includegraphics[width=0.9in]{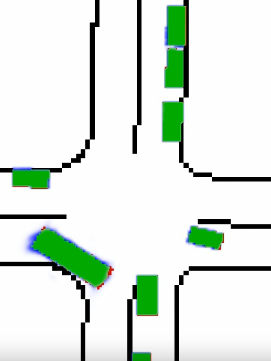}} &
         \subfloat[+2s]{\includegraphics[width=0.9in]{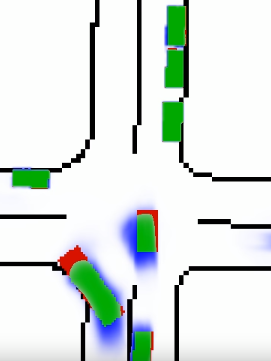}} &
         \subfloat[+4s]{\includegraphics[width=0.9in]{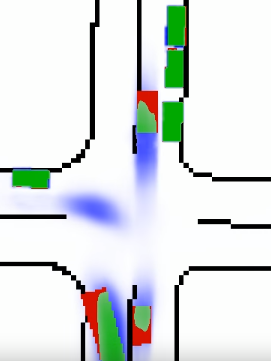}} \\
    \end{tabular}
    \caption{Multi-modal predictions modeled by \networkName{}. In this example, the vehicle in the center is predicted to either continue straight or turn left at the intersection. Images are color coded \textcolor{my-green}{green} $\rightarrow$ true positive (occupancy prediction $>0.5$), \textcolor{blue}{blue} $\rightarrow$ false positive, \textcolor{red}{red} $\rightarrow$ false negative, black $\rightarrow$ rasterized road graph.}
    \label{fig:multimodal}
\end{figure}

\begin{figure}
    \centering
    \begin{tabular}{ccc}
         \subfloat[+0s]{\includegraphics[width=0.9in]{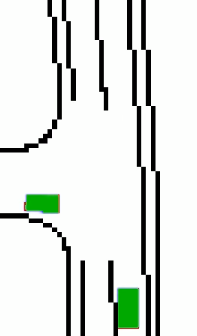}} &
         \subfloat[+2s]{\includegraphics[width=0.9in]{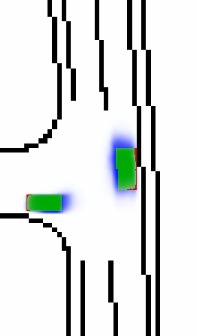}} &
         \subfloat[+4s]{\includegraphics[width=0.9in]{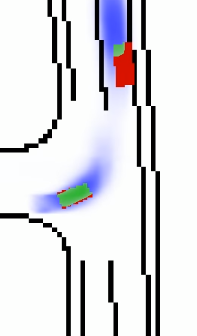}} \\
    \end{tabular}
    \caption{Typical social dynamics are learnt and incorporated in the prediction. In this example, the turning vehicle waits for the vehicle with right of way.}
    \label{fig:socialDynMerge}
\end{figure}

We also qualitatively observe several interesting emergent behaviours of the model\footnote{Samples can be viewed in the supplementary material provided at 
\href{https://sites.google.com/monash.edu/motionperceiver}{https://sites.google.com/monash.edu/motionperceiver}}. $AgentObservation$ performs its function effectively: removing accumulated uncertainty and error in predicted occupancy, and adding unobserved agents to the latent state (Fig. \ref{fig:updatestep}). Agents that are unobserved in the update are preserved in the latent state, but their associated occupancy fades over time. For agents in motion, this manifests as a directed smear, attributed to learned uncertainty estimation over the target's future state. Additionally, \networkName{} produces multi-modal predictions, particularly at decision boundaries for agents (Fig. \ref{fig:multimodal}).
\networkName{} also shows evidence of learning social dynamics, predicting traffic behaviour such as waiting before merging (Fig. \ref{fig:socialDynMerge}).

\subsection{Comparison with Existing Methods} \label{subs:comparison}
For this evaluation, we introduce two modifications to the general baseline model, sampling all past observations and forecasting at the evaluation period. Hence, two $TimePropagate$ modules are used, we refer to this architecture as ``Two Phase". One in the past to propagate $100ms$ to the next observation, another to propagate at $1s$ increments for forecasting. This simple change improves our benchmark score (Two Phase, Table \ref{tab:waymo_abl}) and can indicate that larger $TimePropagate$ increments improves forecasting accuracy, due to less iterations for the same duration. When evaluated on the WOMD withheld test split, our approach demonstrates a significant improvement in Soft IoU compared to existing submissions (Table. \ref{tab:womd_test}). We note that Soft IoU is a potentially more appropriate statistic for forecasting probabilistic occupancy than AUC, particularly due to the multi-modal nature of the task \cite{AUCBad}. Models that are poorly calibrated have significantly worse Soft IoU in Table. \ref{tab:womd_test} than counterparts with similar AUC, a metric that rewards overconfident unimodal predictions.
\begin{table}[hbt]
    \centering
    \caption{Waymo Open Motion Withheld Testing Split Comparison}
    \begin{tabular}{|c|c|c|c|}
        \hline
        \textbf{Model}                  & \textbf{Soft IoU} & \textbf{AUC}  & \textbf{\# Params}\\
        \hline
        Spatial Temporal Convolution    & 0.217             & 0.744         & -\\
        LookAround \cite{LookAround}    & 0.234             & 0.801         & 28.5M\\   
        HOPE \cite{HOPE}                & 0.235             & \textbf{0.803}& 81M\\     
        Temporal Query                  & 0.393             & 0.757         & -\\
        Motionnet                       & 0.411             & 0.694         & -\\
        VectorFlow \cite{VectorFlow}    & 0.488             & 0.755         & 17.1M\\   
        STrajNet \cite{STrajNet}        & 0.491             & 0.778         & 14.5M\\ 
        OFMPNet \cite{OFMPNet}          & 0.502             & 0.770         & 13.3M\\
        YRNet                           & 0.508             & 0.712         & -\\
        \hline
        Ours                            & 0.523             & 0.770         & \textbf{10.7M} \\ 
        Ours + Occ. Flow                & \textbf{0.535}    & 0.779         & \textbf{10.7M} \\
        \hline
    \end{tabular}
    \label{tab:womd_test}
\end{table}

\subsection{Multi-task Training}
The original WOMD challenge included an ``occupancy flow'' task to predict where occupancy has originated from, predominately to recover vehicle identities. The task of predicting the past from the current state runs counter to our architecture that intends to predict the future in a uni-directional stream. However, for fair comparison, we jointly train occupancy flow and obtain marginally improved occupancy results (Ours + Occ. Flow, Table \ref{tab:womd_test}). These marginal gains could potentially be attributed to extra regularization induced by jointly learning a geometrically similar, but distinct task. This is implemented by adding two extra channels to the model output, representing $x$ and $y$ flow, and learned with a simple pixel-wise Huber loss, weighted by $0.1$. We note that a simplified version of occupancy flow ground-truth is used, change in vehicle heading is not considered, the change in agent position is simply broadcast over the flow-mask. With this discrepancy in mind, we obtain a flow end-point-error of $4.900$ on the test split, in line with other models.

\subsection{Runtime Latency}
There is an absence of reporting on the applicability of motion forecasting models for real-time edge inference. Since this is a core benefit of the proposed architecture, we benchmark our system on a common platform, the Nvidia Jetson AGX. To facilitate this, we export \networkName{}'s modules as individual ONNX models to benchmark with trtexec \footnote{\href{https://github.com/NVIDIA/TensorRT/tree/main/samples/trtexec}{https://github.com/NVIDIA/TensorRT/tree/main/samples/trtexec}}. We report results in Table \ref{tab:microbench} with the following parameters: the number of input tokens for signal and agent updates are 16 and 128 respectively, $OccupancyQuery$ generates a $200\times200px$ image, and the rasterized topology input for $RoadgraphEncoder$ is $200\times200px$.
\begin{table}[hbt]
    \centering
    \caption{TensorRT Inference Latency}
    \begin{tabular}{|c|c|}
    \hline
    \textbf{Module}         & \textbf{Inference (ms)}\\
    \hline
    $StateInitialisation$   & 1.202 \\
    $OccupancyQuery$        & 9.260 \\
    $TimePropagate$         & 0.905 \\
    $AgentObservation$      & 0.379 \\
    $SignalObservation$     & 0.460 \\
    $RoadgraphEncoder$      & 0.062 \\
    $RoadContext$           & 0.395 \\
    \hline
    \end{tabular}
    \label{tab:microbench}
\end{table}

Based on these results, our inference latency to forecast $8s$ into the future is $8\cdot(TimePropagate+RoadContext)=9.68ms$, assuming that $TimePropagate$ has been trained to predict $1s$ increments. In parallel, another $TimePropagate$ that matches the scene observation period is applied to anticipate the next update.


\section{Limitations and Further Work}
\networkName{} does not explicitly preserve instance identity in occupancy predictions. This makes the model robust to sensor occlusions and missing information. Although identity is not needed for path planning, future work can explore recovering this by analysing transformer attention. 

Importantly, to be used for path-planning, the model needs to be conditioned on ego-agent actions or those of other agents. A method to achieve this could be to inject updates of potential agent trajectories when forecasting. \networkName{} is lightweight enough that several trajectories can be proposed in parallel. Future work should focus on integrating this with trajectory optimisation strategies that leverage efficient inference with \networkName{}.

\section{Conclusion}
This paper introduces \networkName{}, a motion forecasting architecture designed explicitly for fast and online use. The proposed architecture encodes a scene into a latent state that is evolved forward in time with a learned time evolution function and updated with future observations. This learned recursive state estimation approach is more computationally efficient than existing sequence-based architectures, and performs on-par with larger state-of-the-art models in AUC, and outperforms all others in Soft IoU. Visualized sequences show that \networkName{} is able to learn and incorporate typical social dynamics for prediction such as right-of-way, incorporate observations from multiple sources (eg. traffic lights, road graph), and learn to make realistic probabilistic occupancy predictions.


\bibliographystyle{IEEEtran}
\bibliography{references}

\end{document}